\documentclass[runningheads]{llncs}
\usepackage{graphicx}
\usepackage{comment}
\usepackage{amsmath,amssymb} 
\usepackage{color}
\usepackage{booktabs}

\newcommand{\etal}{\textit{et al}. }

\newcommand{\eg}{\textit{e}.\textit{g}.}

\usepackage{hyperref}
\hypersetup{
  colorlinks, linkcolor=red
}

\usepackage[rightcaption]{sidecap}

\usepackage{graphicx} 
\graphicspath{ {images/} }
\usepackage{wrapfig}

\begin{document}
\pagestyle{headings}
\mainmatter
\def\ECCVSubNumber{116}  

\title{Multiview Detection with\\ Feature Perspective Transformation} 

\titlerunning{Multiview Detection with Feature Perspective Transformation}
%
\author{Yunzhong Hou \and
Liang Zheng \and
Stephen Gould}
\authorrunning{Y. Hou et al.}
%
\institute{
Australian National University \\
Australian Centre for Robotic Vision \\
\email{\{firstname.lastname\}@anu.edu.au}
}
\maketitle

\begin{abstract}
%

Incorporating multiple camera views for detection alleviates the impact of occlusions in crowded scenes. In a multiview detection system, we need to answer two important questions. First, how should we aggregate cues from multiple views? Second, how should we aggregate information from spatially neighboring locations? To address these questions, we introduce a novel multiview detector, MVDet. During multiview aggregation, for each location on the ground, existing methods use multiview anchor box features as representation, which potentially limits performance as pre-defined anchor boxes can be inaccurate. In contrast, via feature map perspective transformation, MVDet employs anchor-free representations with feature vectors directly sampled from corresponding pixels in multiple views. For spatial aggregation, different from previous methods that require design and operations outside of neural networks, MVDet takes a fully convolutional approach with large convolutional kernels on the multiview aggregated feature map. The proposed model is end-to-end learnable and achieves 88.2\% MODA on Wildtrack dataset, outperforming the state-of-the-art by 14.1\%. We also provide detailed analysis of MVDet on a newly introduced synthetic dataset, MultiviewX, which allows us to control the level of occlusion. Code and MultiviewX dataset are available at \url{https://github.com/hou-yz/MVDet}.

\end{abstract}

\section{Introduction}
\label{sec:intro}

Occlusion is a fundamental issue that confronts many computer vision tasks. Specifically, in detection problems, occlusion introduces great difficulties and many methods have been proposed to address it. Some methods focus on the single view detection problem, \eg, part-based detection~\cite{tian2015deep,ouyang2015partial,zhou2017multi}, loss design~\cite{zhang2018occlusion,wang2018repulsion}, and learning non-maximum suppression~\cite{hosang2017learning}. Other methods jointly infer objects from multiple cues, \eg, RGB-D~\cite{gupta2014learning,hoffman2016cross,qi2018frustum}, LIDAR point cloud~\cite{chen2017multi}, and multiple RGB camera views~\cite{fleuret2007multicamera,chavdarova2018wildtrack}. In this paper, we focus on pedestrian detection from multiple RGB camera views (multiview).

\begin{figure}[t]
    \centering
    \includegraphics[width=\linewidth]{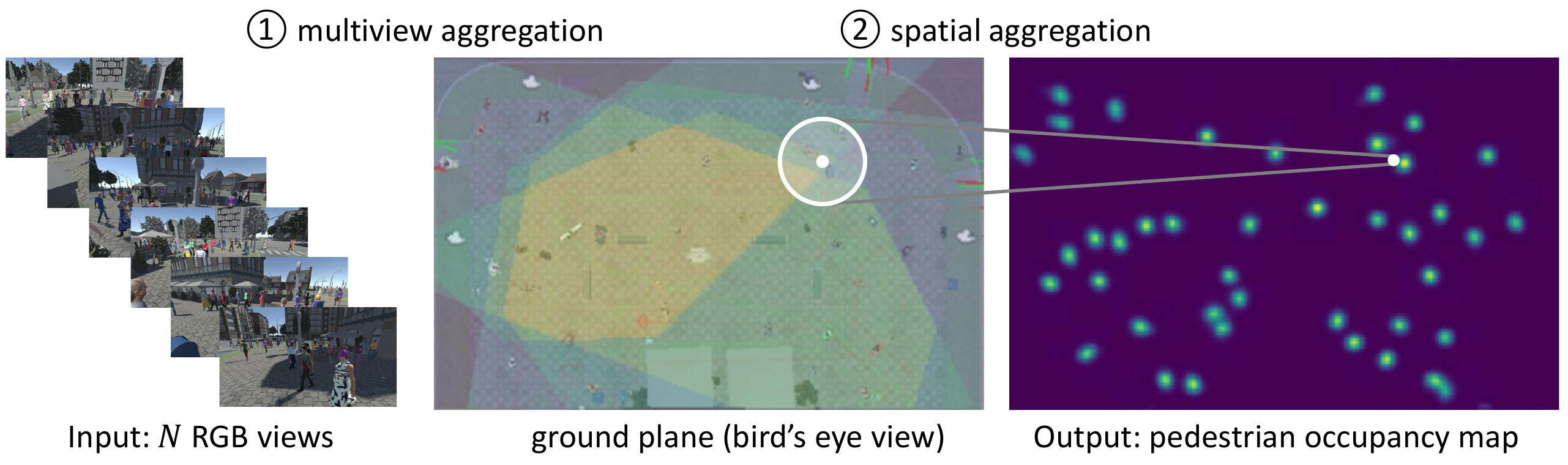}
    \caption{Overview of the multiview pedestrian detection system. \textbf{Left}: the system takes synchronized frames from $N$ cameras as input. \textbf{Middle}: the camera field-of-views overlap on the ground plane, where the multiview cues can be aggregated. \textbf{Right}: the system outputs a pedestrian occupancy map (POM). There are two important questions here. \textbf{First}, how can we aggregate multiple cues. \textbf{Second}, how can we aggregate spatial neighbor information for joint consideration (large white circle), and make a comprehensive decision for pedestrian occupancy (small white circle). 
    }
    \label{fig:system}
\end{figure}

Multiview pedestrian detections usually have synchronized frames from multiple calibrated cameras as input~\cite{fleuret2007multicamera,roig2011conditional,chavdarova2018wildtrack}. These cameras focus on the same area, and have overlapping field-of-view (see Fig.~\ref{fig:system}). 
Camera calibrations provide the matching between 2D image coordinate $\left(u,v\right)$ and 3D world location $\left(x,y,z\right)$. We refer to points with $z=0$ in the 3D world as being on the ground plane (bird's eye view). For each point on the ground plane, based on 3D human width and height assumption, its corresponding bounding box in multiple views can be calculated via projection and then stored. Since the bounding boxes can be retrieved via table lookup, multiview pedestrian detection tasks usually evaluate pedestrian occupancy on the ground plane~\cite{fleuret2007multicamera,chavdarova2018wildtrack}. 

Addressing the ambiguities from occlusions and crowdedness is the main challenge for multiview pedestrian detection. 
Under occlusion, it is difficult to determine if a person exists in a certain location, or how many people exist and where they are. 
To solve this, one must focus on two important aspects of multiview detection: first, \textit{multiview aggregation} and, second, \textit{spatial aggregation} (Fig.~\ref{fig:system}). 
Aggregation of multiview information is essential since having multiple views is the main difference between monocular-view detection and multiview detection. Previously, for a given ground plane location, multiview systems usually choose an anchor-based multiview aggregation approach and represent certain ground plane location with multiview anchor box features~\cite{chavdarova2017deep,baque2017deep,ku2018joint}. However, researchers find the performance of anchor-based methods might be limited by pre-defined anchor boxes in monocular view systems \cite{zhu2019feature,kong2019foveabox,yang2018metaanchor}, while multiview anchor boxes calculated from pre-defined human 3D height and width might also be inaccurate. 
Aggregation of spatial neighbors is also vital for occlusion reasoning. Previous methods \cite{fleuret2007multicamera,roig2011conditional,baque2017deep} usually adopt conditional random field (CRF) or mean-field inference to jointly consider the spatial neighbors. These methods usually requires specific potential terms design or additional operations outside of convolutional neural networks (CNNs). 

In this paper, we propose a simple yet effective method, MVDet, that has heretofore not been explored in the literature for multiview detection. First, for \textit{multiview aggregation}, as representation based on inaccurate anchor boxes can limit system performance, rather than anchor-based approaches~\cite{chavdarova2017deep,baque2017deep,ku2018joint}, MVDet choose an anchor-free representation with feature vectors sampled at corresponding pixels in multiple views. Specifically, MVDet projects the convolution feature map via perspective transformation and concatenates the multiple projected feature maps. 
Second, for \textit{spatial aggregation}, to minimize human design and operations outside of CNN, instead of CRF or mean-field inference \cite{fleuret2007multicamera,roig2011conditional,baque2017deep}, MVDet adopts an fully convolutional solution. 
It applies (learned) convolutions on the aggregated ground plane feature map, and use the large receptive field to jointly consider ground plane neighboring locations. 
The proposed fully convolutional MVDet can be trained in an end-to-end manner. 

We demonstrate the effectiveness of MVDet on two large scale datasets. On Wildtrack, a real-world dataset, MVDet achieved 88.2\% MODA~\cite{kasturi2008framework}, a 14.1\% increase over previous state-of-the-art. On MultiviewX, a synthetic dataset, MVDet also achieves competitive results under multiple levels of occlusions.

\section{Related Work}
\label{sec:related}

\textbf{Monocular view detection.} 
Detection is one of the most important problems in computer vision. 
Anchor-based methods like faster R-CNN~\cite{ren2015faster} and SSD~\cite{liu2016ssd} achieve great performance. Recently, finding pre-defined anchors might limit performance, many anchor-free methods are proposed~\cite{zhu2019feature,tian2019fcos,kong2019foveabox,yang2018metaanchor,duan2019centernet,law2018cornernet}.
On pedestrian detection, 
some researchers detect pedestrian bounding boxes through head-foot point detection~\cite{song2018small} or center and scale detection~\cite{liu2019high}. Occlusion handling in pedestrian detection draws great attention from the research community. 
Part-based detectors are very popular~\cite{ouyang2015partial,tian2015deep,noh2018improving,zhang2018occlusion} since the occluded people are only partially observable. 
Hosang \etal \cite{hosang2017learning} learn non-maximal suppression for occluded pedestrians. 
Repulsion loss~\cite{wang2018repulsion} is proposed to repulse bounding boxes. 

\textbf{3D object understanding with multiple information sources.} 
Incorporating multiple information sources, such as depth, point cloud, and other RGB camera views is studied for 3D object understanding. 
For multiple view 3D object classification, Su \etal \cite{su2015multi} use maximum pooling to aggregate the features from different 2D views. 
For 3D object detection, aggregating information from RGB image and LIDAR point cloud are widely studied. Chen \etal \cite{chen20153d} investigate 3D object detection with stereo image. 
View aggregation for 3D anchors is studied in \cite{ku2018joint}, where the researchers extract features for every 3D anchor from RGB camera and LIDAR bird's eye view. Liang \etal \cite{liang2018deep} calculate the feature for each point from bird's eye view as multi-layer perceptron output from camera view features of $K$ nearest neighbor LIDAR points. Frustum PointNets~\cite{qi2018frustum} first generate 2D bounding boxes proposal from RGB image, then extrude them to 3D viewing frustums. 
Yao \etal edit attributes of 3D vehicle models to create content consistent vehicle dataset \cite{yao2019simulating}.

\textbf{Multiview pedestrian detection.} 
In multiview pedestrian detections, first, aggregating information from multiple RGB cameras is essential. In \cite{chavdarova2017deep,baque2017deep}, researchers fuse multiple information source for multiview 2D anchors. Given fixed assumption of human width and height, all ground plane locations and their corresponding multiview 2D anchor boxes are first calculated. Then, researchers in \cite{chavdarova2017deep,baque2017deep} represent ground plane position with corresponding anchor box features. In \cite{fleuret2007multicamera,xu2016multi,roig2011conditional}, single view detection results are fused instead. 
Second, in order to aggregate spatial neighbor information, mean-field inference~\cite{fleuret2007multicamera,baque2017deep} and conditional random field (CRF)~\cite{roig2011conditional,baque2017deep} are exploited. 
In \cite{fleuret2007multicamera,baque2017deep}, the overall occupancy in the scenario is cast as an energy minimization problem and solved with CRF. 
Fleuret \etal ~\cite{fleuret2007multicamera} first estimate ideal 2D images under certain occupancy, and then compare them with the real multiview inputs. 
Baque \etal \cite{baque2017deep} construct higher-order potentials as consistency between CNN estimations and generated ideal images, and train the CRF with CNN in a combined manner, and achieve state-of-the-art performance on Wildtrack dataset~\cite{chavdarova2018wildtrack}.

\textbf{Geometric transformation in deep learning.} Geometric transformations such as affine transformation and perspective transformation can model many phenomena in computer vision, and can be explicitly calculated with a fixed set of parameters. Jaderberg \etal \cite{jaderberg2015spatial} propose Spatial Transformer Network that learns the affine transformation parameters for translation and rotation on the 2D RGB input image. Wu \etal \cite{wu2016single} estimate the projection parameters and project 2D key points from the 3D skeleton. Yan \etal \cite{yan2016perspective} translate one 3D volume to 2D silhouette via perspective transformation. Geometry-aware scene text detection is studied in \cite{wang2018geometry} through estimating instance-level affine transformation. For cross-view image retrieval, Shi \etal \cite{shi2019spatial} apply polar transformation to bring the representations closer in feature space. 
Lv \etal propose a perspective-aware generative model for novel view synthesis for vehicles \cite{lv2020pose}. 



\begin{figure}[t]
    \centering
    \includegraphics[width=\linewidth]{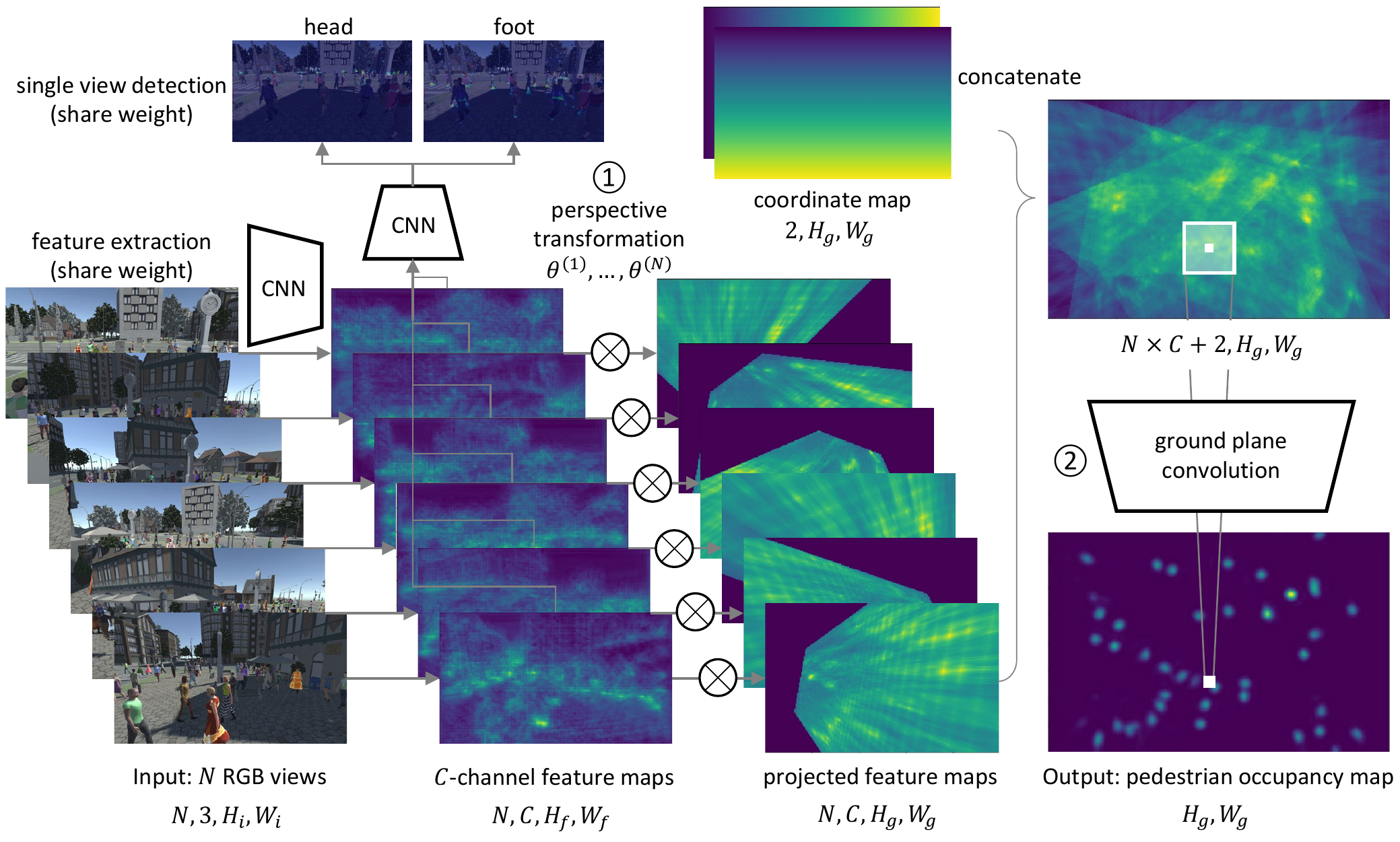}
    \caption{MVDet architecture. First, given input images of shape $\left[3,H_i,W_i\right]$ from $N$ cameras, the proposed network uses a CNN to extract $C$-channel feature maps for each input image. The CNN feature extractor here shares weight among $N$ inputs. Next, we reshape the $C$-channel feature maps into the size of $\left[H_f,W_f\right]$, and run single view detection by detecting the head-foot pairs. Then, for \textbf{multiview aggregation} (circled 1), we take an anchor-free approach and combine the perspective transformation of $N$ feature maps based on their camera calibrations $\theta^{\left(1\right)}, \dots, \theta^{\left(N\right)}$, which results in $N$ feature maps of shape $\left[C,H_g,W_g\right]$. For each ground plane location, we store its X-Y coordinates in a $2$-channel coordinate map~\cite{liu2018intriguing}. Through concatenating $N$ projected feature maps with a coordinate map, we aggregate the ground plane feature map for the whole scenario (of shape $\left[N \times C + 2,H_g,W_g\right]$). At last, we apply large kernel convolutions on the ground plane feature map, so as to \textbf{aggregate spatial neighbor information} (circled 2) for a joint and comprehensive final occupancy decision. 
    }
    \label{fig:architecture}
\end{figure}

\section{Methodology}
\label{sec:method}
In this work, we focus on the occluded pedestrian detection problem in an multiview scenario and design MVDet for dealing with ambiguities. 
MVDet features anchor-free \textit{multiview aggregation} that alleviate the influence from inaccurate anchor boxes in previous works \cite{chen2017multi,ku2018joint,chavdarova2017deep,baque2017deep}, and fully convolutional \textit{spatial aggregation} that does not rely on CRF or mean-field inference~\cite{fleuret2007multicamera,roig2011conditional,baque2017deep}. 
As shown in Fig.~\ref{fig:architecture}, MVDet takes multiple RGB images as input, and outputs the pedestrian occupancy map (POM) estimation. In the following sections, we will introduce the proposed multiview aggregation (Section~\ref{sec:sec:aggreagtion}), spatial aggregation  (Section \ref{sec:sec:joint}), and training and testing configurations (Section \ref{sec:sec:training}).

\subsection{Multiview Aggregation}
\label{sec:sec:aggreagtion}

Multiview aggregation is a very important part of multiview systems. In this section, we explain the anchor-free aggregation method in MVDet that alleviate influence from inaccurate anchor boxes, and compare it with several alternatives. 

\textbf{Feature map extraction. }
In MVDet, first, given $N$ images of shape $\left[H_i,W_i\right]$ as input ($H_i$ and $W_i$ denote the image height and width), the proposed architecture uses a CNN to extract $N$ $C$-channel feature maps (Fig.~\ref{fig:architecture}). Here, we choose ResNet-18~\cite{he2016deep} for its strong performance and light-weight. This CNN calculates $C$-channel feature maps separately for $N$ input images, while sharing weight among all calculations. In order to maintain a relatively high spatial resolution for the feature maps, we replace the last 3 strided convolutions with dilated convolutions~\cite{yu2015multi}. 
Before projection, we resize $N$ feature maps into a fixed size $\left[H_f,W_f\right]$ ($H_f$ and $W_f$ denote the feature map height and width). 
In each view, similar to \cite{law2018cornernet,song2018small}, we then detect pedestrians as a pair of head-foot points with a shared weight single view detector.

\begin{figure}[t]
    \centering
    \includegraphics[width=\linewidth]{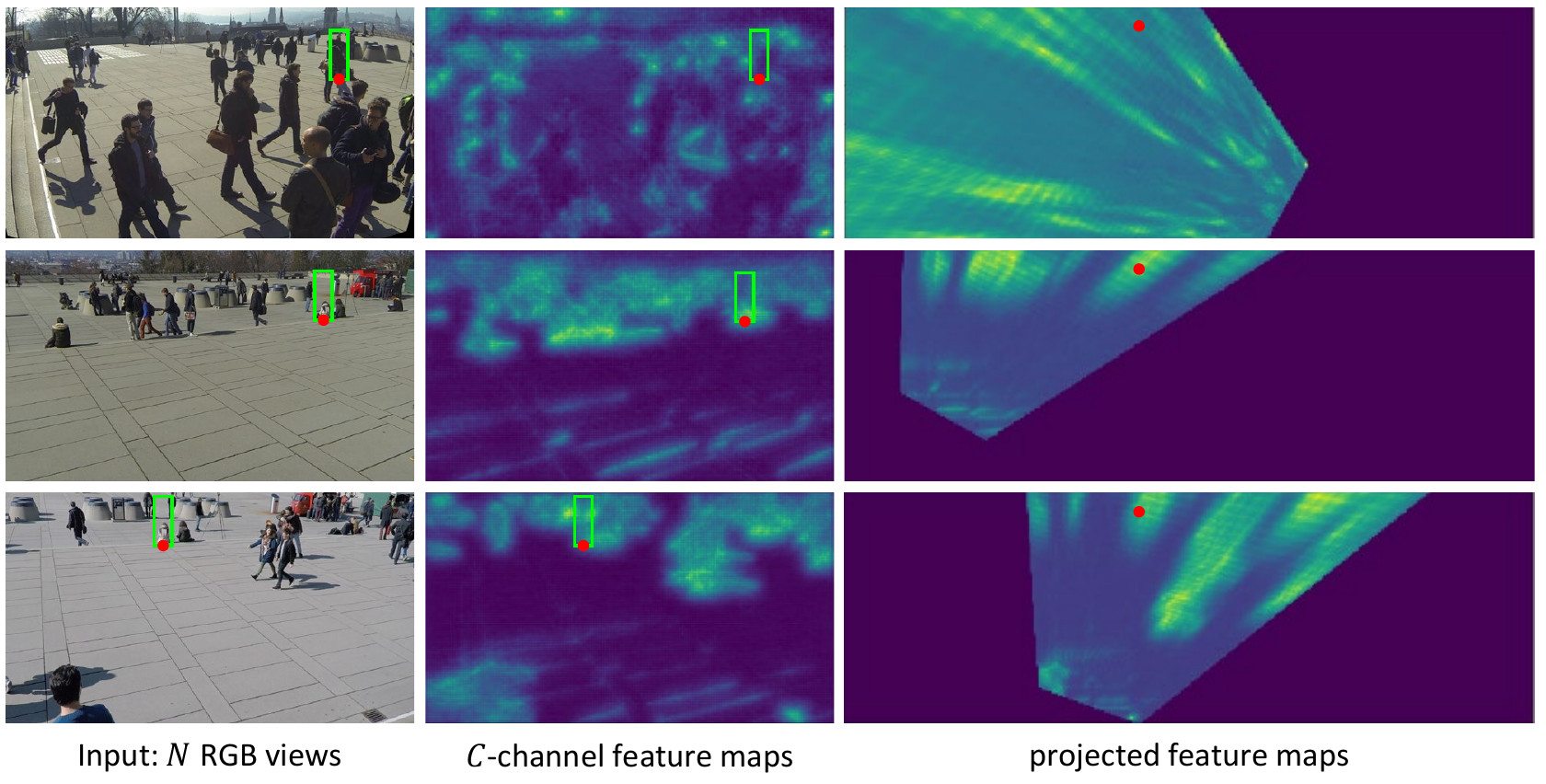}
    \caption{Representing ground plane locations with feature vectors (anchor-free) or anchor boxes features (anchor-based). Red dots represent a certain ground plane location and its corresponding pixel in different views. Green bounding boxes refer to anchor boxes corresponding to that ground plane location. As human targets might not be the same size as assumed (\eg, sitting lady in white coat), ROI-pooling for multiview anchor boxes might fail to provide the most accurate feature representation for that location. On the contrary, being anchor-free, feature vectors retrieved from corresponding points avoids the impact of inaccurate anchor boxes. 
    }
    \label{fig:bbox_vs_projection}
\end{figure}

\textbf{Anchor-free representation.} 
Previously, in detection tasks that have multiple cues, \eg, 3D object detection and multiview pedestrian detection, anchor-based representation is commonly adopted \cite{chen2017multi,ku2018joint,chavdarova2017deep,baque2017deep}. 
Specifically, one can represent a ground plane location (red points in Fig.~\ref{fig:bbox_vs_projection}) with anchor box (green boxes in Fig.~\ref{fig:bbox_vs_projection}) features via ROI-pooling \cite{girshick2015fast}. As the size and shape of anchor boxes are calculated from assumed 3D human height and width \cite{chavdarova2017deep,baque2017deep}, 
these anchor boxes might \textit{not} be accurate, which potentially limits system performance~\cite{zhu2019feature,kong2019foveabox,yang2018metaanchor}. As in Fig.~\ref{fig:bbox_vs_projection}, the lady in the white coat is sitting and only takes up half of the anchor box. As a result, ROI pooling will result in feature representation that describes the background to a large extent and causes confusion. 

In contrast, being anchor-free, the proposed method represents ground plane locations with feature vectors sampled from feature maps at corresponding points, which avoids the impact of inaccurate anchor boxes. Given camera calibrations, the corresponding points can be retrieved accurately. 
With learnable convolution kernels, these feature vectors can represent information from an adaptive region in its receptive field. 
As a result, ground plane feature maps constructed via anchor-free feature representation avoid pooling from inaccurate anchor boxes, and still contains sufficient information from 2D images for detection.

\begin{figure}[t]
    \centering
    \includegraphics[width=\linewidth]{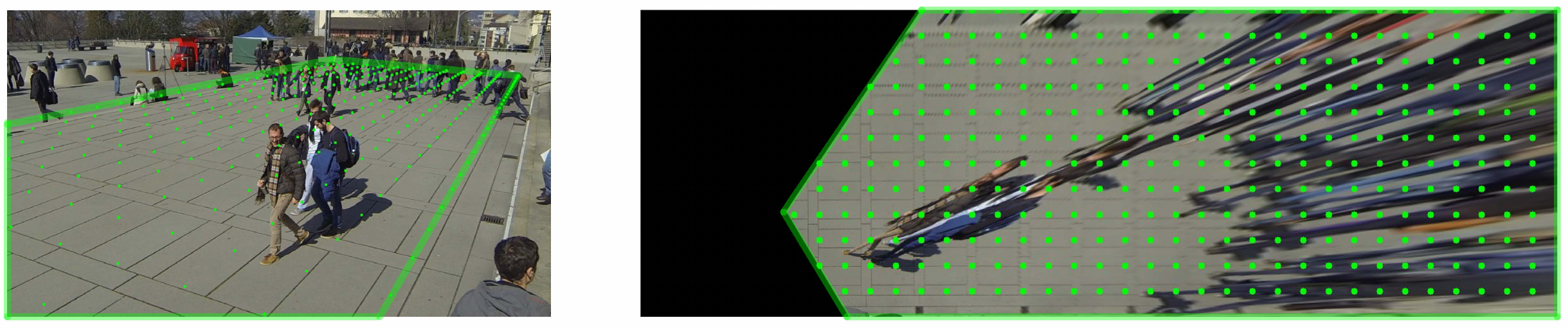}
    \caption{Illustration of perspective transformation. Assuming all pixels are on the ground plane ($z=0$), we can use a parameterized sampling grid (green dots) to project a 2D image (left) to the ground plane (right). The remaining locations are padded with $0$.
    }
    \label{fig:perspective_trans}
\end{figure}

\textbf{Perspective transformation.} 
To retrieve anchor-free representations, we project feature maps with perspective transformation. Translation between 3D locations $\left(x,y,z\right)$ and 2D image pixel coordinates $\left(u,v\right)$ is done via
\begin{equation}
\label{eq:perspective_3x4}
    s\left(\begin{matrix}u \\ v \\ 1\end{matrix}\right) = P_\theta \left(\begin{matrix} x \\ y \\ z \\ 1\end{matrix}\right) = A \left[R|\mathbf{t}\right] \left(\begin{matrix} x \\ y \\ z \\ 1\end{matrix}\right) = \left[\begin{matrix} \theta_{11} & \theta_{12} & \theta_{13} & \theta_{14} \\ \theta_{21} & \theta_{22} & \theta_{23} & \theta_{24} \\ \theta_{31} & \theta_{32} & \theta_{33} & \theta_{34} \end{matrix}\right] \left(\begin{matrix} x \\ y \\ z \\ 1\end{matrix}\right),
\end{equation}
where $s$ is a real-valued scaling factor, and $P_\theta$ is a $3\times4$ perspective transformation matrix. Specifically, $A$ is the $3\times3$ intrinsic parameter matrix. 
$\left[R|\textbf{t}\right]$ is the $3\times4$ joint rotation-translation matrix, or extrinsic parameter matrix, where $R$ specifies the rotation and $\mathbf{t}$ specifies the translation. 

A point (pixel) from an image lies on a line in the 3D world. To determine exact 3D locations of image pixels, we consider a common reference plane: the ground plane, $z=0$. For all 3D location $\left(x,y,0\right)$ on the ground plane, the point-wise transformation can be written as
\begin{equation}
\label{eq:perspective_3x3}
    s\left(\begin{matrix}u \\ v \\ 1\end{matrix}\right) = P_{\theta,0} \left(\begin{matrix} x \\ y \\ 1\end{matrix}\right) = \left[\begin{matrix} \theta_{11} & \theta_{12} & \theta_{14} \\ \theta_{21} & \theta_{22} & \theta_{24} \\ \theta_{31} & \theta_{32} & \theta_{34} \end{matrix}\right] \left(\begin{matrix} x \\ y \\ 1\end{matrix}\right),
\end{equation}
where $P_{\theta,0}$ denotes the $3\times3$ perspective transformation matrix that have the third column canceled from $P_{\theta}$. 

To implement this within neural networks, we quantize the ground plane locations into a grid of shape $\left[H_g,W_g\right]$. For camera $n\in \left\{1,\dots,N\right\}$ with calibration $\theta^{\left(n\right)}$, we can project the image onto the $z=0$ ground plane by applying a parameterized sampling grid of shape $\left[H_g,W_g\right]$ based on Equation~\ref{eq:perspective_3x3}. These sampling grids generate projected feature maps on the ground plane, where the remaining (out-of-view) locations are padded with zero (Fig.~\ref{fig:perspective_trans}). 
We concatenate a 2-channel coordinate map \cite{liu2018intriguing} to specify the X-Y coordinates for ground plane locations (Fig.~\ref{fig:architecture}). Together with projected $C$-channel feature maps from $N$ cameras, we have a $\left(N\times C+2\right)$ channel ground plane feature map.

\begin{figure}[t]
    \centering
    \includegraphics[width=\linewidth]{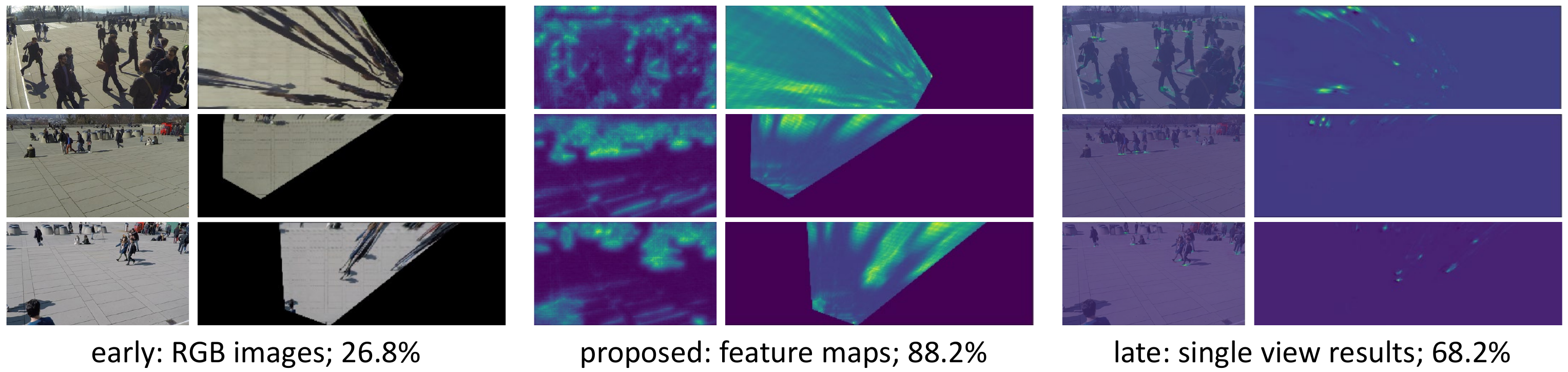}
    \caption{Different projection choices and their performances (MODA~\cite{kasturi2008framework}). \textbf{Left}: early projection of RGB images breaks the spatial relationship between RGB pixels, which introduces great difficulties to the convolutions. \textbf{Right}: late projection of single view detection results (foot) limits the information to be aggregated. \textbf{Middle}: the proposed projection of feature maps not only are more robust to the pixel structure break (high-level semantic feature can represent information by itself, thus suffer less from structure break), but also contain more information. 
    }
    \label{fig:aggregation_stage}
\end{figure}

\textbf{Different projection choices. }
For multiview aggregation, there are multiple choices for projection: we can project the RGB images, feature maps, or single view results (Fig.~\ref{fig:aggregation_stage}). 
First, RGB pixels on its own contains relatively little information, and much information is preserved in the spatial structures. However, projection breaks the spatial relationship between neighboring RGB pixels. As a result, this limits the performance of the multiview detector. 
Second, projecting the single view results (detected foot points) limits the information to be aggregated. In fact, in this setup, the system has no access to cues other than the single view detection results. Since single view results might not be accurate under occlusion (which is the reason for introducing multiple views), this setup can also limit the overall performance. 
In this paper, we propose to project the feature map. Compared to other choices, feature maps not only suffer less from the spatial structure break (since 2D spatial information have already been concentrated into individual pixels in feature maps), but also contain more information. As shown in Fig.~\ref{fig:aggregation_stage}, aggregation via feature maps projection achieves highest MODA~\cite{kasturi2008framework} performance.

\subsection{Spatial aggregation}
\label{sec:sec:joint}
In the previous section, we show that multiview information can be aggregated in an anchor-free manner through perspective transformation and concatenation. One remaining problem is how to aggregate information from spatial neighbors. 

Occlusion are generated by human crowd within a certain area. To deal with the ambiguities, one can consider the certain area and the human crowd in that area jointly for an overall informed decision. 
Previously, CRFs and mean-field inference are adopted, but requires design and operations besides CNN. In this work, we propose an fully convolutional alternative with large kernel convolutions on the ground plane feature map. In fact, Zheng \etal \cite{zheng2015conditional} find that CNN can model some behavior and characteristics of CRFs. And Peng \etal \cite{peng2017large} outperform CRFs with large kernel convolutions for semantic segmentation. 
In MVDet, we feed the $\left(N\times C+2\right)$ channel ground plane feature map to convolution layers that have a relatively large receptive field, so as to jointly consider the ground plane neighbors. 
Here, we use three layers of dilated convolution for having minimal parameters while still keeping a larger ground plane receptive field. 
The last layer outputs an $1$-channel $\left[H_g,W_g\right]$ pedestrian occupancy map (POM) $\mathbf{\Tilde{g}}$ with no activation.

\subsection{Training and Testing}
\label{sec:sec:training}
\textbf{In training}, we train MVDet as a regression problem. Given ground truth pedestrian occupancy $\mathbf{g}$, similar to landmark detection \cite{cao2017realtime}, we use a Gaussian kernel $f\left(\cdot\right)$ to generate a ``\textit{soft}'' ground truth target $f\left(\mathbf{g}\right)$. In order to train the whole network, we use Euclidean distance $\left\|\cdot\right\|_2$ between network output $\mathbf{\Tilde{g}}$ and ``\textit{soft}'' target $f\left(\mathbf{g}\right)$ as loss function,
\begin{equation}
    \label{eq:ground_loss}
    \mathcal{L}_{\text{ground}}=\left\|\mathbf{\Tilde{g}}-f\left(\mathbf{g}\right)\right\|_2. 
\end{equation}
We also include bounding box regression loss from $N$ camera inputs as another supervision. The single view head-foot detection is also trained as a regression problem. For single view detection results $\mathbf{\Tilde{s}}_{\text{head}}^{\left(n\right)},\mathbf{\Tilde{s}}_{\text{foot}}^{\left(n\right)}$ and the corresponding ground truth $\mathbf{s}_{\text{head}}^{\left(n\right)},\mathbf{s}_{\text{foot}}^{\left(n\right)}$ in view $n\in\left\{1,...,N\right\}$, the loss is computed as,
\begin{equation}
    \label{eq:single_view_loss}
    \mathcal{L}_{\text{single}}^{\left(n\right)}
    =
    \left\|\mathbf{\Tilde{s}}_{\text{head}}^{\left(n\right)}-f\left(\mathbf{s}_{\text{head}}^{\left(n\right)}\right)\right\|_2
    +
    \left\|\mathbf{\Tilde{s}}_{\text{foot}}^{\left(n\right)}-f\left(\mathbf{s}_{\text{foot}}^{\left(n\right)}\right)\right\|_2. 
\end{equation}
Combining ground plane loss $\mathcal{L}_{\text{ground}}$ and $N$ single view losses $\mathcal{L}_{\text{single}}^{\left(n\right)}$, we have the overall loss for training MVDet,
\begin{equation}
    \label{eq:combined_loss}
    \mathcal{L}_{\text{combined}}=\mathcal{L}_{\text{ground}}+\alpha \times \frac{1}{N}\sum_{n=1}^{N}{\mathcal{L}_{\text{single}}^{\left(n\right)}}, 
\end{equation}
where $\alpha$ is a hyper-parameter for singe view loss weight.

\textbf{During testing}, MVDet outputs a single-channel occupancy probability map $\mathbf{\Tilde{g}}$. We filter the occupancy map with a minimum probability of $0.4$, and then apply non-maximum suppression (NMS) on the proposals. This NMS uses a Euclidean distance threshold of $0.5$ meters, which is the same threshold for considering this location proposal as true positive in evaluation \cite{chavdarova2018wildtrack}.

\section{Experiment}
\label{sec:experiment}

\subsection{Experiment Setup}
\label{sec:sec:datasets}

\textbf{Datasets.} We test on two multiview pedestrian detection datasets (Table~\ref{tab:datasets}). 

The \textit{Wildtrack} dataset includes 400 synchronized frames from 7 cameras, covering a 12 meters by 36 meters region. For annotation, the ground plane is quantized into a $480\times 1440$ grid, where each grid cell is a 2.5-centimeter square. The 7 cameras capture images with a $1080\times 1920$ resolution, and are annotated at 2 frames per second (fps). On average, there are 20 persons per frame in Wildtrack dataset and each locations in the scene is covered by 3.74 cameras.

The \textit{MultiviewX} dataset is a new synthetic dataset collected for multiview pedestrian detection. 
We use Unity engine~\cite{unity} to create the scenario. As for pedestrians, we use human models from PersonX~\cite{sun2019dissecting}. MultiviewX dataset covers a slightly smaller area of 16 meters by 25 meters. Using the same 2.5-centimeter square grid cell, we quantize the ground plane into a $640\times 1000$ grid. There are 6 cameras with overlapping field-of-view in MultiviewX dataset, each of which outputs a $1080\times 1920$ resolution image. We also generate annotations for 400 frames in MultiviewX at 2 fps (same as Wildtrack). On average, 4.41 cameras cover the same location. 
Being a synthetic dataset, there are various potential configurations for the scenario with free annotations. 
Under the default setting, MultiviewX has 40 persons per frame, doubling the crowdedness in Wildtrack. 
If not specified, MultiviewX refers to this default setting.

\begin{table}[t]
\centering
\caption{Datasets comparison for multiview pedestrian detection}
\begin{tabular}{l|c|c|c|c|c|c}
\toprule
\multicolumn{1}{c|}{} & \#camera & resolution & frames & area  & crowdedness & avg. coverage\\ \hline
Wildtrack  & 7                             & $1080\times 1920$                    & 400                         & $12\times 36$ m$^2$                                    & 20 person/frame  & 3.74 cameras                          \\ \hline
MultiviewX & 6                             & $1080\times 1920$                       & 400                         & $16\times 25$ m$^2$                                  & 40 person/frame  & 4.41 cameras                           \\ 
\bottomrule
\end{tabular}
\label{tab:datasets}
\end{table}

\textbf{Evaluation metrics. }
Following \cite{chavdarova2018wildtrack}, we use the first $90\%$ frames in both datasets for training, and the last $10\%$ frames for testing. 
We report precision, recall, MODA, and MODP. MODP evaluates the localization precision, whereas MODA accounts for both the false positives and false negatives~\cite{kasturi2008framework}. We use MODA as the primary performance indicator, as it considers both false positives and false negatives. A threshold of $0.5$ meters is used to determine true positives. 

\subsection{Implementation Details}
\label{sec:sec:implementation}
For memory usage concerns, we downsample the $1080\times 1920$ RGB images to $H_i=720, W_i=1280$. We remove the last two layers (global average pooling; classification output) in ResNet-18~\cite{he2016deep} for $C=512$ channel feature extraction, and use dilated convolution to replace the strided convolution. This results in a $8\times$ downsample from the $720\times1280$ input. Before projection, we bilinearly interpolate the feature maps into shape of $H_f=270, W_f=480$. With $4\times$ down sample, for Wildtrack and MultiviewX, the ground plane grid sizes are set as $H_g=120,W_g=360$ and $H_g=160,W_g=250$, respectively, where each cell represents a 10 centimeter square. For spatial aggregation, we use 3 convolutional layers with $3\times 3$ kernels and dilation of $1,2,4$. This will increase the receptive field for each ground plane location (cell) to $15\times 15$ square cells, or $1.5\times 1.5$ square meters. 
In order to train MVDet, we use an SGD optimizer with a momentum of 0.5, L2-normalization of $5\times 10^{-4}$. The weight $\alpha$ for single view loss is set to 1. We use the one-cycle learning rate scheduler~\cite{smith2019super} with the max learning rate set to 0.1, and train for 10 epochs with batch size set to 1. We finish all experiments on two RTX-2080Ti GPUs. 

\begin{table}[t]
\centering
\caption{Multiview aggregation and spatial aggregation in different methods}
\begin{tabular}{l|l|l}
\toprule
Method                           & Multiview aggregation                  & Spatial aggregation \\ \hline
RCNN \& clustering~\cite{xu2016multi}               & detection results & clustering                       \\ \hline
POM-CNN~\cite{fleuret2007multicamera}                          & detection results & mean-field inference                 \\ \hline
DeepMCD~\cite{chavdarova2017deep}                          & anchor box features        & N/A                              \\ \hline
Deep-Occlusion~\cite{baque2017deep}                   & anchor box features        &  CRF + mean-field inference          \\ \hline\hline
MVDet (project images)        & RGB image pixels             & large kernel convolution                    \\ \hline
MVDet (project results)       & detection results & large kernel convolution                      \\ \hline
MVDet (w/o large kernel)         & feature maps      & N/A                              \\ \hline
\textbf{MVDet}          & feature maps      & large kernel convolution                     \\
\bottomrule
\end{tabular}
\label{tab:notations}
\end{table}

\subsection{Method Comparisons}
\label{sec:sec:variants}

In Table~\ref{tab:notations}, we compare multiview aggregation and spatial aggregation in different methods. 
For multiview aggregation, previous methods either project single view detection results \cite{xu2016multi,fleuret2007multicamera} or use multiview anchor box features \cite{chavdarova2017deep,baque2017deep}. For spatial aggregation, clustering \cite{xu2016multi}, mean-field inference \cite{fleuret2007multicamera,baque2017deep}, and CRF \cite{baque2017deep,roig2011conditional} are investigated. 
In order to compare against previous methods, we create the following variants for MVDet. 
To compare anchor-free aggregation with anchor-based methods, we create ``MVDet (w/o large kernel)'', which remove the large kernel convolutions. 
This variant is created as a direct comparison against DeepMCD~\cite{chavdarova2017deep}, both of which do not include spatial aggregation. 
To compare different projection choices (Section~\ref{sec:sec:aggreagtion}), we include two variants that either project RGB image pixels ``MVDet (project images)'' or single view detection results ``MVDet (project results)''. 
``MVDet (w/o large kernel)'' also show the effectiveness of spatial aggregation. 
All variants follow the same training protocol as original MVDet.

\begin{table}[t]
\centering
\caption{Performance comparison with state-of-the-art methods on multiview pedestrian detection datasets. $^*$ indicates that the results are from our implementation }
\begin{tabular}{l||c|c|c|c||c|c|c|c}
\toprule
                        & \multicolumn{4}{c||}{Wildtrack}                              & \multicolumn{4}{c}{MultiviewX}                               \\ \hline
Method                  & MODA          & MODP          & Prec.   & Recall        & MODA          & MODP          & Prec.     & Recall        \\ \hline
RCNN \& clustering      & 11.3          & 18.4          & 68          & 43            & 18.7$^*$          & 46.4$^*$          & 63.5$^*$          & 43.9$^*$          \\ \hline
POM-CNN                 & 23.2          & 30.5          & 75          & 55            & -             & -             & -             & -             \\ \hline
DeepMCD                 & 67.8          & 64.2          & 85          & 82            & 70.0$^*$          & 73.0$^*$          & 85.7$^*$          & 83.3$^*$          \\ \hline
Deep-Occlusion          & 74.1          & 53.8          & \textbf{95} & 80            & 75.2$^*$          & 54.7$^*$          & \textbf{97.8}$^*$ & 80.2$^*$          \\ \hline\hline
MVDet (project images)        & 26.8          & 45.6          & 84.2        & 33.0          & 19.5          & 51.0          & 84.4          & 24.0          \\ \hline
MVDet (project results)       & 68.2          & 71.9          & 85.9        & 81.2          & 73.2          & 79.7          & 87.6          & 85.0          \\ \hline
MVDet (w/o large kernel) & 76.9          & 71.6          & 84.5        & 93.5          & 77.2          & 76.3          & 89.5          & 85.9 \\  \hline
\textbf{MVDet} & \textbf{88.2} & \textbf{75.7} & 94.7        & \textbf{93.6} & \textbf{83.9} & \textbf{79.6} & 96.8          & \textbf{86.7}          \\
\bottomrule
\end{tabular}
\label{tab:sota}
\end{table}

\subsection{Evaluation of MVDet}
\label{sec:sec:evaluation}

\textbf{Comparison against state-of-the-art methods.}
In Table~\ref{tab:sota}, we compare the performance of MVDet against multiple state-of-the-art methods on multiview pedestrian detection. Since there are no available codes for some of the methods, for a fair comparison on MultiviewX, we re-implement these methods to the best as we can. 
On Wildtrack dataset, MVDet achieves 88.2\% MODA, a +14.1\% increase over previous state-of-the-art. On MultiviewX dataset, MVDet achieves 83.9\% MODA, an 8.7\% increase over our implementation of Deep-Occlusion~\cite{baque2017deep}. 
MVDet also achieves highest MODP and recall on both datasets, but slightly falls behind Deep-Occlusion in terms of precision. 
It is worth mentioning that Deep-Occlusion outperforms MVDet in terms of precision, but falls behind in terms of recall. This shows that their CNN-CRF method is very good at suppressing the false positives, but sometimes has a tendency to miss a few targets. 

\textbf{Effectiveness of anchor-free multiview aggregation.} 
Even without spatial aggregation, ``MVDet (w/o large kernel)'' achieves 76.9\% MODA on Wildtrack dataset and 77.2\% MODA on MultiviewX dataset. In fact, it slightly outperforms current state-of-the-art by +2.8\% and +2.0\% on two datasets. 
The high performance proves the effectiveness of our anchor-free aggregation via feature map projection. 
In Section~\ref{sec:sec:aggreagtion}, we hypothesize that inaccurate anchor boxes could possibly result in less accurate aggregated features and thus proposed an anchor-free approach. In Table~\ref{tab:sota}, we prove the effectiveness of our anchor-free approach by comparing anchor-based DeepMCD \cite{chavdarova2017deep} against anchor-free ``MVDet (w/o large kernel)'', both of which do not include spatial aggregation. The variant of MVDet outperforms DeepMCD by 9.1\% on Wildtrack dataset, and 7.2\% MODA on MultiviewX dataset, which demonstrates anchor-free feature maps projection can be a better choice for multiview aggregation in multiview pedestrian detection when the anchor boxes are not accurate. 

Feature map projection brings less improvement over multiview anchor box features on MultiviewX dataset (+7.2\% on MultiviewX compared to +9.1\% on Wildtrack). This is because MultiviewX dataset has synthetic humans, whereas Wildtrack captures real-world pedestrians. Naturally, the variances of human height and width are higher in the real-world scenario, as synthetic humans are of very similar sizes. This suggests less accurate anchor boxes on average for the real-world dataset, Wildtrack. As a result, aggregation via feature map projection brings larger improvement on Wildtrack dataset.

\begin{figure}[t]
    
    \includegraphics[width=\linewidth]{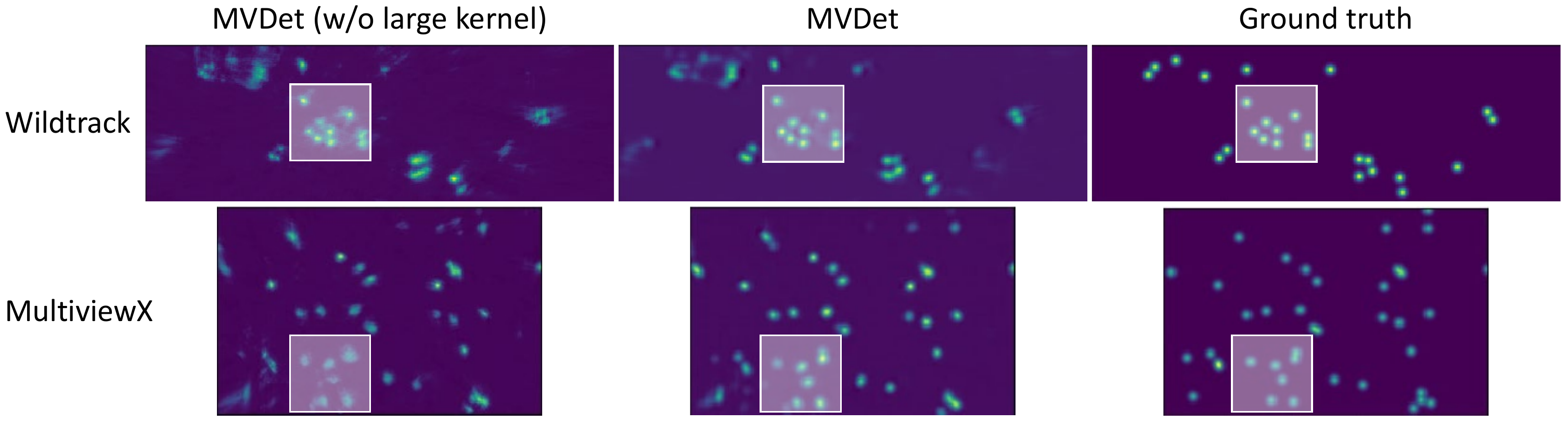}
    \caption{Effectiveness of spatial aggregation via large kernel convolution. Compared to ``MVDet (w/o large kernel)'', MVDet outputs occupancy probabilities more similar to the ground truth, especially in highlighted areas.
    }
    \label{fig:spatial_aggregation}
\end{figure}

\textbf{Comparison between different projection choices.} 
We claim that projecting the feature maps is a better choice than projecting the RGB images or single view results in Section~\ref{sec:sec:aggreagtion}. 
Projecting the RGB images breaks the spatial relationship between pixels, and a single RGB pixel represents little information. As a result, in Table~\ref{tab:sota}, we find ``MVDet (project images)'' leads to largely inferior performance on both datasets (26.8\% and 19.5\% MODA). 
Although single view results are robust to spatial patter break from projection, the information contained in them is limited. Due to crowdedness and occlusion, single view detection might lose many true positives. As such, clustering these projected single view results as in ``RCNN \& clustering''~\cite{xu2016multi} are proven to be extremely difficult (11.3\% and 18.7\% MODA). 
Replacing the clustering with large kernel convolution ``MVDet (project results)'' increases the performance by a large margin (68.2\% and 73.2\% MODA), as it alleviates the problem of formulating 1-size clusters (clusters that have only one component, as the detections are missing from occlusion) and can be trained in an end-to-end manner. Still, the restricted information in detection results prevents the variant from higher performance.

\textbf{Effectiveness of spatial aggregation via large kernel convolution.} 
Spatial aggregation with large kernel convolutions brings forward a +11.3\% MODA increase on Wildtrack dataset, and a +6.7\% performance increase on MultiviewX dataset. In comparison, spatial aggregation with CRF and mean-field inference brings forward increases of +6.3\% and +5.2\% on the two datasets, going from DeepMCD to Deep-Occlusion. 
We do not assert superiority of either the CRF-based or CNN-based methods. 
We only argue that the proposed CNN-based method can effectively aggregate spatial neighbor information to address the ambiguities from crowdedness or occlusion while need no design or operations besides CNN. As shown in Fig.~\ref{fig:spatial_aggregation}, large kernel convolutions manages to output results that are more similar to the ground truth. 

For spatial aggregation, both the proposed large kernel convolution and CRF bring less improvement on MultiviewX dataset. 
As mentioned in Table~\ref{tab:datasets}, even though there are fewer cameras in MultiviewX dataset, 
each ground plane location in MultiviewX dataset is covered by more cameras on average. Each location is covered by 4.41 cameras (field-of-view) on average for MultiviewX dataset, as opposed to 3.74 in Wildtrack. More camera coverage usually introduces more information and reduces the ambiguities, which also limits the performance increase from addressing ambiguities via spatial aggregation. 

\begin{figure}[t]
    \includegraphics[width=\linewidth]{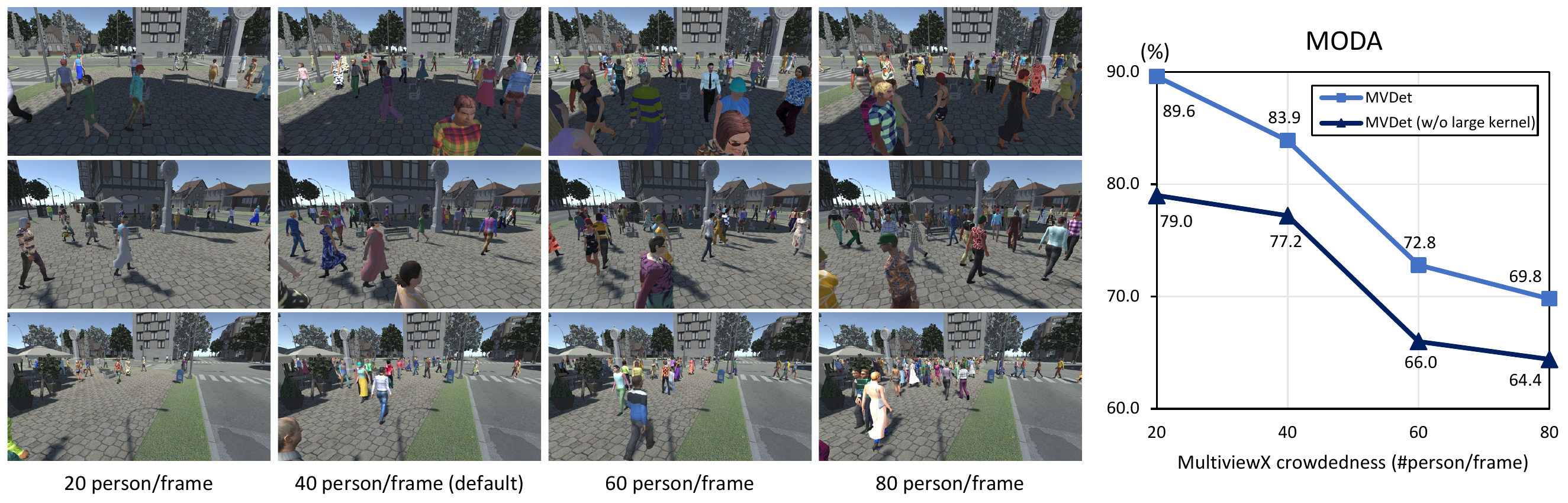}
    \caption{MultiviewX dataset under different crowdedness configuration (left), and corresponding MVDet performance (right). 
    }
    \label{fig:crowdedness}
\end{figure}

\textbf{Influence of different crowdedness and occlusion levels. }
Being a synthetic dataset, there are multiple available configurations for MultiviewX. In Fig. \ref{fig:crowdedness} (left), we show the camera views under multiple levels of crowdedness. As the crowdedness of the scenario increases, the occlusion also increases. In Fig. \ref{fig:crowdedness} (right), we show the MVDet performance under multiple levels of occlusions. As crowdedness and occlusions increase (more difficult), MODA of both MVDet and MVDet ``MVDet (w/o large kernel)'' decrease. In addition, performance increases from spatial aggregation also drop, due to the task being more challenging and heavy occlusion also affecting the spatial neighbors.

\begin{wrapfigure}{l}{0.4\textwidth}
    \centering
    \includegraphics[width=\linewidth]{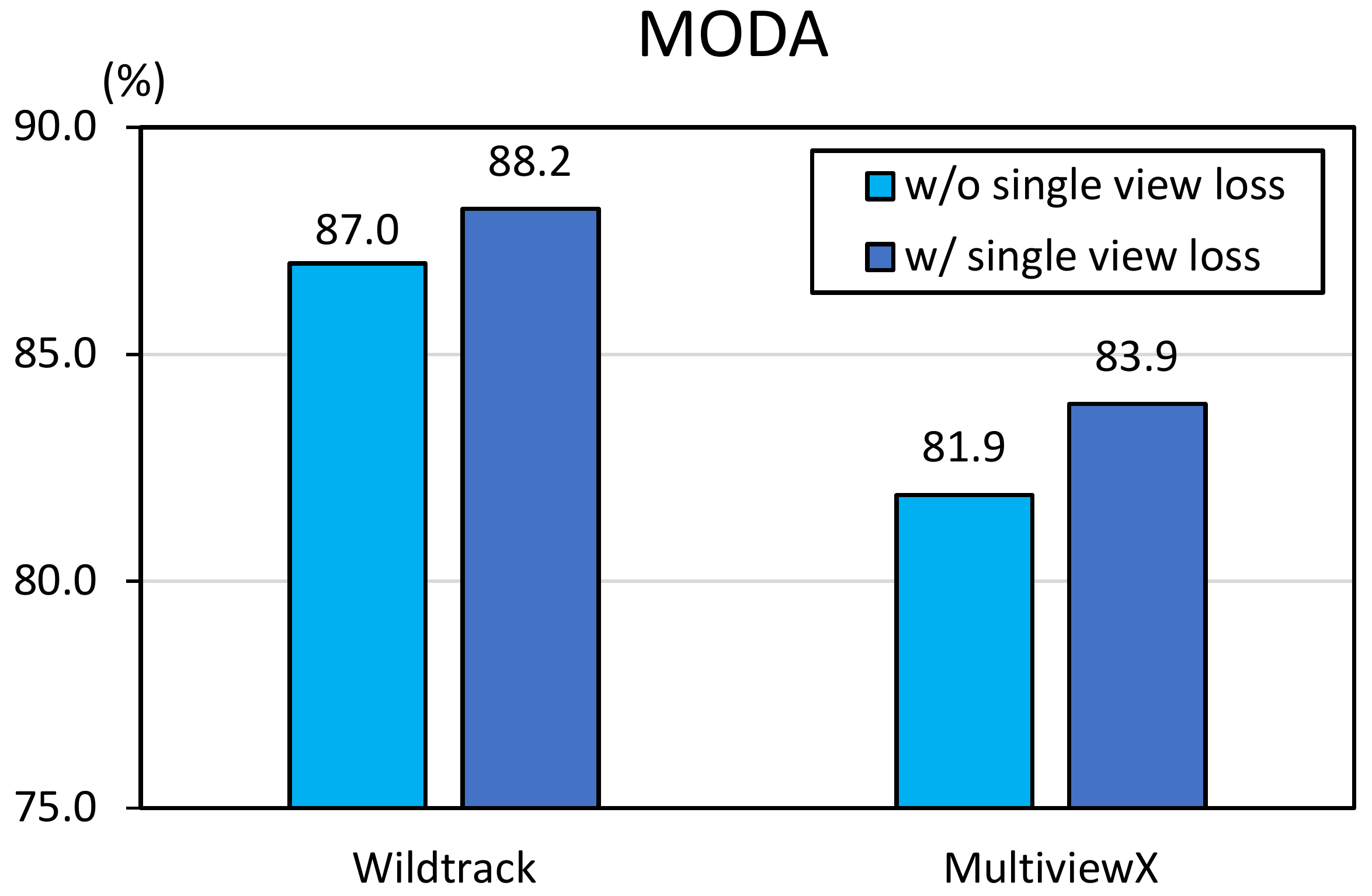}
    \caption{MODA performance of MVDet with ($\alpha=1$) or without ($\alpha=0$) single view detection loss. 
    }
    \label{fig:single_view_loss}
\end{wrapfigure}

\textbf{Influence of single view detection loss.} 
In our default setting of MVDet, for the combined loss in Eq.~\ref{eq:combined_loss}, the ratio $\alpha$ is set to 1. 
In Fig.~\ref{fig:single_view_loss}, we investigate the influence of removing the single view loss. 
Without single view detection loss, we find a -1.2\% and a -2.0\% performance loss on both datasets, which are still very competitive. In fact, we believe single view foot detection loss does not further benefit the system, as the foot points are already supervised on the ground plane. The head point detection loss, on the other hand, can produce heterogeneous supervision, thus further improving system performance. As discussed in Section~\ref{sec:sec:aggreagtion} and Section~\ref{sec:sec:evaluation}, less accurate bounding box annotations limit the performance gain from single view loss on Wildtrack dataset.

\section{Conclusion}
In this paper, we investigate pedestrian detection in a crowded scene, through incorporating multiple camera views. Specifically, we focus on addressing the ambiguities that arise from occlusion with multiview aggregation and spatial aggregation, two core aspects of multiview pedestrian detection. For multiview aggregation, rather than anchor-based approach in previous methods, we take an anchor-free approach by combining the projected feature maps. For spatial aggregation, different from previous methods that need design and operations aside from CNN, we apply large kernels in our fully convolutional approach. 
The proposed system, MVDet, achieves 88.2\% MODA on Wildtrack dataset, outperforming previous state-of-the-art by 14.1\%. On MultiviewX, a new synthetic dataset for multiview pedestrian detection, MVDet also achieves very competitive results. We believe the proposed MVDet can serve as a strong baseline for multiview pedestrian detection, encouraging further studies in related fields.

\section*{Acknowledgement}
Dr. Liang Zheng is the recipient of Australian Research Council Discovery Early Career Award (DE200101283) funded by the Australian Government. The authors thank all anonymous reviewers and ACs for their constructive comments.

%
%
\bibliographystyle{splncs04}
\bibliography{egbib}

\begin{thebibliography}{10}
\providecommand{\url}[1]{\texttt{#1}}
\providecommand{\urlprefix}{URL }
\providecommand{\doi}[1]{https://doi.org/#1}

\bibitem{baque2017deep}
Baqu{\'e}, P., Fleuret, F., Fua, P.: Deep occlusion reasoning for multi-camera
  multi-target detection. In: Proceedings of the IEEE International Conference
  on Computer Vision. pp. 271--279 (2017)

\bibitem{cao2017realtime}
Cao, Z., Simon, T., Wei, S.E., Sheikh, Y.: Realtime multi-person 2d pose
  estimation using part affinity fields. In: Proceedings of the IEEE conference
  on computer vision and pattern recognition. pp. 7291--7299 (2017)

\bibitem{chavdarova2018wildtrack}
Chavdarova, T., Baqu{\'e}, P., Bouquet, S., Maksai, A., Jose, C., Bagautdinov,
  T., Lettry, L., Fua, P., Van~Gool, L., Fleuret, F.: Wildtrack: A multi-camera
  hd dataset for dense unscripted pedestrian detection. In: Proceedings of the
  IEEE Conference on Computer Vision and Pattern Recognition. pp. 5030--5039
  (2018)

\bibitem{chavdarova2017deep}
Chavdarova, T., et~al.: Deep multi-camera people detection. In: 2017 16th IEEE
  International Conference on Machine Learning and Applications (ICMLA). pp.
  848--853. IEEE (2017)

\bibitem{chen20153d}
Chen, X., Kundu, K., Zhu, Y., Berneshawi, A.G., Ma, H., Fidler, S., Urtasun,
  R.: 3d object proposals for accurate object class detection. In: Advances in
  Neural Information Processing Systems. pp. 424--432 (2015)

\bibitem{chen2017multi}
Chen, X., Ma, H., Wan, J., Li, B., Xia, T.: Multi-view 3d object detection
  network for autonomous driving. In: Proceedings of the IEEE Conference on
  Computer Vision and Pattern Recognition. pp. 1907--1915 (2017)

\bibitem{duan2019centernet}
Duan, K., Bai, S., Xie, L., Qi, H., Huang, Q., Tian, Q.: Centernet: Keypoint
  triplets for object detection. In: Proceedings of the IEEE International
  Conference on Computer Vision. pp. 6569--6578 (2019)

\bibitem{fleuret2007multicamera}
Fleuret, F., Berclaz, J., Lengagne, R., Fua, P.: Multicamera people tracking
  with a probabilistic occupancy map. IEEE transactions on pattern analysis and
  machine intelligence  \textbf{30}(2),  267--282 (2007)

\bibitem{girshick2015fast}
Girshick, R.: Fast r-cnn object detection with caffe. Microsoft Research
  (2015)

\bibitem{gupta2014learning}
Gupta, S., Girshick, R., Arbel{\'a}ez, P., Malik, J.: Learning rich features
  from rgb-d images for object detection and segmentation. In: European
  conference on computer vision. pp. 345--360. Springer (2014)

\bibitem{he2016deep}
He, K., Zhang, X., Ren, S., Sun, J.: Deep residual learning for image
  recognition. In: Proceedings of the IEEE conference on computer vision and
  pattern recognition. pp. 770--778 (2016)

\bibitem{hoffman2016cross}
Hoffman, J., Gupta, S., Leong, J., Guadarrama, S., Darrell, T.: Cross-modal
  adaptation for rgb-d detection. In: 2016 IEEE International Conference on
  Robotics and Automation (ICRA). pp. 5032--5039. IEEE (2016)

\bibitem{hosang2017learning}
Hosang, J., Benenson, R., Schiele, B.: Learning non-maximum suppression. In:
  Proceedings of the IEEE conference on computer vision and pattern
  recognition. pp. 4507--4515 (2017)

\bibitem{jaderberg2015spatial}
Jaderberg, M., Simonyan, K., Zisserman, A., et~al.: Spatial transformer
  networks. In: Advances in neural information processing systems. pp.
  2017--2025 (2015)

\bibitem{kasturi2008framework}
Kasturi, R., Goldgof, D., Soundararajan, P., Manohar, V., Garofolo, J., Bowers,
  R., Boonstra, M., Korzhova, V., Zhang, J.: Framework for performance
  evaluation of face, text, and vehicle detection and tracking in video: Data,
  metrics, and protocol. IEEE Transactions on Pattern Analysis and Machine
  Intelligence  \textbf{31}(2),  319--336 (2008)

\bibitem{kong2019foveabox}
Kong, T., Sun, F., Liu, H., Jiang, Y., Shi, J.: Foveabox: Beyond anchor-based
  object detector. arXiv preprint arXiv:1904.03797  (2019)

\bibitem{ku2018joint}
Ku, J., Mozifian, M., Lee, J., Harakeh, A., Waslander, S.L.: Joint 3d proposal
  generation and object detection from view aggregation. In: 2018 IEEE/RSJ
  International Conference on Intelligent Robots and Systems (IROS). pp.~1--8.
  IEEE (2018)

\bibitem{law2018cornernet}
Law, H., Deng, J.: Cornernet: Detecting objects as paired keypoints. In:
  Proceedings of the European Conference on Computer Vision (ECCV). pp.
  734--750 (2018)

\bibitem{liang2018deep}
Liang, M., Yang, B., Wang, S., Urtasun, R.: Deep continuous fusion for
  multi-sensor 3d object detection. In: Proceedings of the European Conference
  on Computer Vision (ECCV). pp. 641--656 (2018)

\bibitem{liu2018intriguing}
Liu, R., Lehman, J., Molino, P., Such, F.P., Frank, E., Sergeev, A., Yosinski,
  J.: An intriguing failing of convolutional neural networks and the coordconv
  solution. In: Advances in Neural Information Processing Systems. pp.
  9605--9616 (2018)

\bibitem{liu2016ssd}
Liu, W., Anguelov, D., Erhan, D., Szegedy, C., Reed, S., Fu, C.Y., Berg, A.C.:
  Ssd: Single shot multibox detector. In: European conference on computer
  vision. pp. 21--37. Springer (2016)

\bibitem{liu2019high}
Liu, W., Liao, S., Ren, W., Hu, W., Yu, Y.: High-level semantic feature
  detection: A new perspective for pedestrian detection. In: Proceedings of the
  IEEE Conference on Computer Vision and Pattern Recognition. pp. 5187--5196
  (2019)

\bibitem{lv2020pose}
Lv, K., Sheng, H., Xiong, Z., Li, W., Zheng, L.: Pose-based view synthesis for
  vehicles: A perspective aware method. IEEE Transactions on Image Processing
  \textbf{29},  5163--5174 (2020)

\bibitem{noh2018improving}
Noh, J., Lee, S., Kim, B., Kim, G.: Improving occlusion and hard negative
  handling for single-stage pedestrian detectors. In: Proceedings of the IEEE
  Conference on Computer Vision and Pattern Recognition. pp. 966--974 (2018)

\bibitem{ouyang2015partial}
Ouyang, W., Zeng, X., Wang, X.: Partial occlusion handling in pedestrian
  detection with a deep model. IEEE Transactions on Circuits and Systems for
  Video Technology  \textbf{26}(11),  2123--2137 (2015)

\bibitem{peng2017large}
Peng, C., Zhang, X., Yu, G., Luo, G., Sun, J.: Large kernel matters--improve
  semantic segmentation by global convolutional network. In: Proceedings of the
  IEEE conference on computer vision and pattern recognition. pp. 4353--4361
  (2017)

\bibitem{qi2018frustum}
Qi, C.R., Liu, W., Wu, C., Su, H., Guibas, L.J.: Frustum pointnets for 3d
  object detection from rgb-d data. In: Proceedings of the IEEE Conference on
  Computer Vision and Pattern Recognition. pp. 918--927 (2018)

\bibitem{ren2015faster}
Ren, S., He, K., Girshick, R., Sun, J.: Faster r-cnn: Towards real-time object
  detection with region proposal networks. In: Advances in neural information
  processing systems. pp. 91--99 (2015)

\bibitem{roig2011conditional}
Roig, G., Boix, X., Shitrit, H.B., Fua, P.: Conditional random fields for
  multi-camera object detection. In: 2011 International Conference on Computer
  Vision. pp. 563--570. IEEE (2011)

\bibitem{shi2019spatial}
Shi, Y., Liu, L., Yu, X., Li, H.: Spatial-aware feature aggregation for image
  based cross-view geo-localization. In: Advances in Neural Information
  Processing Systems. pp. 10090--10100 (2019)

\bibitem{smith2019super}
Smith, L.N., Topin, N.: Super-convergence: Very fast training of neural
  networks using large learning rates. In: Artificial Intelligence and Machine
  Learning for Multi-Domain Operations Applications. vol. 11006, p. 1100612.
  International Society for Optics and Photonics (2019)

\bibitem{song2018small}
Song, T., Sun, L., Xie, D., Sun, H., Pu, S.: Small-scale pedestrian detection
  based on topological line localization and temporal feature aggregation. In:
  Proceedings of the European Conference on Computer Vision (ECCV). pp.
  536--551 (2018)

\bibitem{su2015multi}
Su, H., Maji, S., Kalogerakis, E., Learned-Miller, E.: Multi-view convolutional
  neural networks for 3d shape recognition. In: Proceedings of the IEEE
  international conference on computer vision. pp. 945--953 (2015)

\bibitem{sun2019dissecting}
Sun, X., Zheng, L.: Dissecting person re-identification from the viewpoint of
  viewpoint. In: Proceedings of the IEEE Conference on Computer Vision and
  Pattern Recognition. pp. 608--617 (2019)

\bibitem{tian2015deep}
Tian, Y., Luo, P., Wang, X., Tang, X.: Deep learning strong parts for
  pedestrian detection. In: Proceedings of the IEEE international conference on
  computer vision. pp. 1904--1912 (2015)

\bibitem{tian2019fcos}
Tian, Z., Shen, C., Chen, H., He, T.: Fcos: Fully convolutional one-stage
  object detection. In: Proceedings of the IEEE international conference on
  computer vision. pp. 9627--9636 (2019)

\bibitem{unity}
Unity: Unity technologies, \url{https://unity.com/}

\bibitem{wang2018geometry}
Wang, F., Zhao, L., Li, X., Wang, X., Tao, D.: Geometry-aware scene text
  detection with instance transformation network. In: Proceedings of the IEEE
  Conference on Computer Vision and Pattern Recognition. pp. 1381--1389 (2018)

\bibitem{wang2018repulsion}
Wang, X., Xiao, T., Jiang, Y., Shao, S., Sun, J., Shen, C.: Repulsion loss:
  Detecting pedestrians in a crowd. In: Proceedings of the IEEE Conference on
  Computer Vision and Pattern Recognition. pp. 7774--7783 (2018)

\bibitem{wu2016single}
Wu, J., Xue, T., Lim, J.J., Tian, Y., Tenenbaum, J.B., Torralba, A., Freeman,
  W.T.: Single image 3d interpreter network. In: European Conference on
  Computer Vision. pp. 365--382. Springer (2016)

\bibitem{xu2016multi}
Xu, Y., Liu, X., Liu, Y., Zhu, S.C.: Multi-view people tracking via
  hierarchical trajectory composition. In: Proceedings of the IEEE Conference
  on Computer Vision and Pattern Recognition. pp. 4256--4265 (2016)

\bibitem{yan2016perspective}
Yan, X., Yang, J., Yumer, E., Guo, Y., Lee, H.: Perspective transformer nets:
  Learning single-view 3d object reconstruction without 3d supervision. In:
  Advances in neural information processing systems. pp. 1696--1704 (2016)

\bibitem{yang2018metaanchor}
Yang, T., Zhang, X., Li, Z., Zhang, W., Sun, J.: Metaanchor: Learning to detect
  objects with customized anchors. In: Advances in Neural Information
  Processing Systems. pp. 320--330 (2018)

\bibitem{yao2019simulating}
Yao, Y., Zheng, L., Yang, X., Naphade, M., Gedeon, T.: Simulating content
  consistent vehicle datasets with attribute descent. arXiv preprint
  arXiv:1912.08855  (2019)

\bibitem{yu2015multi}
Yu, F., Koltun, V.: Multi-scale context aggregation by dilated convolutions.
  arXiv preprint arXiv:1511.07122  (2015)

\bibitem{zhang2018occlusion}
Zhang, S., Wen, L., Bian, X., Lei, Z., Li, S.Z.: Occlusion-aware r-cnn:
  detecting pedestrians in a crowd. In: Proceedings of the European Conference
  on Computer Vision (ECCV). pp. 637--653 (2018)

\bibitem{zheng2015conditional}
Zheng, S., Jayasumana, S., Romera-Paredes, B., Vineet, V., Su, Z., Du, D.,
  Huang, C., Torr, P.H.: Conditional random fields as recurrent neural
  networks. In: Proceedings of the IEEE international conference on computer
  vision. pp. 1529--1537 (2015)

\bibitem{zhou2017multi}
Zhou, C., Yuan, J.: Multi-label learning of part detectors for heavily occluded
  pedestrian detection. In: Proceedings of the IEEE International Conference on
  Computer Vision. pp. 3486--3495 (2017)

\bibitem{zhu2019feature}
Zhu, C., He, Y., Savvides, M.: Feature selective anchor-free module for
  single-shot object detection. In: Proceedings of the IEEE Conference on
  Computer Vision and Pattern Recognition. pp. 840--849 (2019)

\end{thebibliography}
\end{document}